# Subject Specific Stream Classification Preprocessing Algorithm for Twitter Data Stream

N. H. N. D. de Silva, M. K. D. T. Maldeniya and D. N. C. Wijeratne

Department of Computer Science and Engineering, University of Moratuwa, Sri Lanka

*Abstract*— **Micro-blogging service Twitter is a lucrative source for data mining applications on global sentiment. But due to the omnifariousness of the subjects mentioned in each data item; it is inefficient to run a data mining algorithm on the raw data. This paper discusses an algorithm to accurately classify the entire stream in to a given number of mutually exclusive collectively exhaustive streams upon each of which the data mining algorithm can be run separately yielding more relevant results with a high efficiency.**

## I. INTRODUCTION

TWITTER is a social networking and micro-blogging website which allows registered users to send messages called tweets. A user can follow other users in order to receive their tweets. If the user account is not set to the option private, the tweets from that account are also obtainable from the public timeline. The public timeline and the tweets of the users that a certain user is following can be accessed using the Twitter API. A tweet is a text-based post which has an upper limit of 140 characters. Usually people tweet about their personal life and their opinion of the world and its events. With 190 million users worldwide, twitter is a valuable data source to gain on-the-fly information about the current situation of the world.

Because of the open-ended nature of Twitter, the tweets are spread across all the aspects of human life. This fact considered with the rate of which the tweets are added to the twitter stream, it is not very effective to run a data mining algorithm on the raw twitter stream.

The objective of this paper is to introduce an algorithm to select the best set of word attributes to classify the twitter stream for in a program which intends to run a data mining algorithm upon the twitter stream. The algorithm described in this paper will break the incoming twitter stream in to mutually exclusive collectively exhaustive classes upon each of which the data mining algorithm can be run to identify traits present within each class in relation to the class itself and time. Using the mined data it is possible to draw connections between the classes. The data mining tool Weka, developed by University of Waikato [1] was used to measure the accuracy of the algorithm.

## II. DESIGN AND IMPLEMENTATION

Though twitter is a continuous data stream and data mining tool Weka is built for static data mining purposes, the overall design assumes for blocks of tweets to be available. Thus data preprocessing is carried out separately from the twitter stream. Since the objective of the research was to develop a suitable data preprocessing methodology to be used in data mining tools such as Weka, this approach was decided to be acceptable.

The overall design and operation of the data preprocessing methodology consists of 5 distinct stages, after which a data model is built through Weka data mining tool.

### A. Data Collecting Methodology

Data required to train the data model were collected through the tweet collecting desktop and online application "tHE ARCHIVISt" [2], [3]. The public tweets that were available on the five categories art, lifestyle, politics, technology and business were collected during the periods of $24^{th}$ November to $11^{th}$ December, 2010 and $17^{th}$ to $25^{th}$ February, 2011. The data collected by "tHE ARCHIVISt" application were mostly through twitter search facilities thus the accuracy of the data obtained was significantly high.

Data were put into separate files according to the class to which the tweets were categorized. Each data file was created containing around 11,000 tweets with the class name of it as the filename. Around 300,000 tweets were collected with varying composition on the 5 classes.

### B. Indexing Algorithm

Next step was to select suitable attributes to index the collected tweets. It was found that there are two ways to do this;

1. Build the list of attributes using the words occurring in the already classified tweets

2. Use a dictionary of words and use them as attributes depending on their significance in classifying a tweet in to the class that it is classified to.

It was found that although the first approach would give attributes with direct correlation to the collected tweets, it comes with some inherent problems. The most severe problem was filtering out tweets that are not in English. Although languages using Cyrillic, Chinese, Arabic, etc. scripts could be easily filtered out, languages such as French, German, Spanish, etc. which use the Latin script to write was indistinguishable from English owing to the fact that English too is written using the Latin script. The typical behavior of the twitter users posed a problem too. Since there is a 140 character limit for a tweet people tend to use pseudo-words and on some accounts drop punctuation marks and spaces.

Because of the above problems the second approach was selected. For this end it was decided to use the WordNet[4] lexical database created by Cognitive Science Laboratory of Princeton University, since it is free and contains a well categorized list of over 150,000 different words. The first approach to be used is to list the words in each of the 45 WordNet files under the name of the file and use the file names as the attributes. Appendix A contains an extract from the "noun.person" file. Thus in this approach the words; Zeus, Hera, Greek_deity, deity, God were listed under the attribute "noun.person"

When the index was queried with a tweet; it was programmed to go through the words of the tweet while looking up each word in the said 45 lists. An integer array with a length of 45 was maintained and for each occurs of a word in a give list; the matching integer value was increased. Finally the integer array was returned as the output.

Although this approach was sound for tweet sets of the order of millions, for the number of tweets that we were going to use in this project the attributes were too coarse-grained. Thus it was decided to extend the indexing mechanism. Owing to the fact that WordNet lexical database itself is not a flat file, most of the words were mentioned with a hypernym. Thus it was possible to group words with the same hypernym together and set those words under an attribute by the name of the hypernym. For an example form Appendix A it is evident that the words "Zeus" and "Hera" have "Greek_deity" as the hypernym and in return "Greek_deity" has "deity" as the hypernym. These chains of inheritance were extracted from WordNet. Then all the words that came as a hypernym for one or more word was removed from the basic word list and was added to a separate list. This hypernym list contained 11551 words. Words without a hypernym were listed using the old file name approach. For an example the word "self" in Appendix A was listed under "noun.person". Then chain inheritance was considered and words in the base list was linked to the top most ancestor and all the nodes in between were dropped. This resulted in an attribute count of 7906 plus the 45 filename attributes. Then a threshold function was introduced to filter the hypernyms based on the number of words that came as hyponyms of it. The words under the hypernyms that got disqualified at this stage were reclassified under the filename algorithm. It was found that the optimum value for the threshold value for the hypernym influence to be 2500 in which case an attribute count of 262 was returned inclusive of the 45 filenames. The querying mechanism was not needed to be altered after this change in indexing algorithm.

*C. Recursive Duplicate Elimination Algorithm*

Elimination of duplicate data from the trained data set is a vital component in achieving higher accuracy in the trained model. This is carried out under three stages.

1) *Removal of duplicate tweets:* This eliminates occurrence of the same tweet in the data set over and over again. Even if two tweets are different, due to substitution of the usernames and URLs with "user" and "url" words respectively, if other content of the tweet is the same, it would be considered as the same tweet.

2) *Removal of unclassified data items:* Due to limitations of WordNet structure, certain tweets' words will not be categorized in to any of the attributes. Thus such data items are eliminated as they are insignificant for the data model.

3) *Recursive Duplicate Elimination Algorithm:* The Comma Separated Value (csv) data items generated by extracting data from tweets and using the indexing algorithm consisted of many duplicates. In some cases same data was available for multiple classes reducing the efficiency and accuracy of the trained model. A recursive duplicate elimination algorithm was devised to remove such discrepancies Fig. 1 and 2.

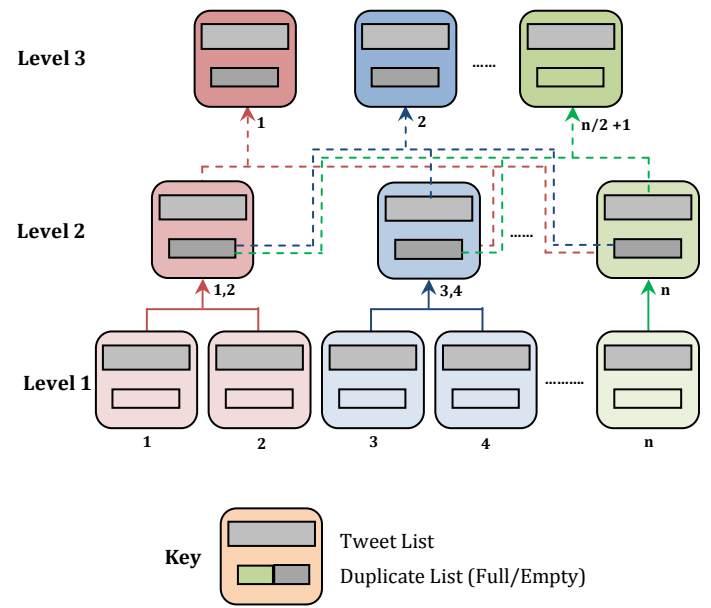

Fig. 1. Recursive duplicate elimination algorithm (No of data files assumed to be odd for the illustration purposes)

Objects are created for each data file with the tweet list and its duplicate tweet list. The algorithm was designed to incorporate in to threads in order to improve the utilization of multi-core capabilities of the computer and to increase the speed of execution.

**RECURSIVE-DUPLICATE-ELIMINATION( )**

```
while duplicateExist
  do A.add(All Data Files)
    for i ← 0 to length[A]/2
      do tempObj ← REMOVE-INTER-CLASS-DUPLICATES(A[i],A[length[A] – (i+1)])
         B.add(tempObj)

    if length[A]%2 > 1  ▷ if no of files is odd
      then B.add(A[length[A]/2])

    for i ← 0 to length[B]
      do tempObj ← B[i]
        for j ← 0 to length[B]
          do if i ≠ j
            then tempObj ← REMOVE-CLASS-DUPLICATES(tempObj,B[j])
         C.add(tempobj)

    if length[C] = 1
      then duplicateExist ← false
    else
      then A ← C
```

Fig. 2. Recursive duplicate elimination algorithm

The REMOVE-INTER-CLASS-DUPLICATES method eliminates duplicates between two data files and stores the duplicates in the resulting file for the use of others. REMOVE-CLASS-DUPLICATES eliminate the duplicate processed tweets in a data object by comparing another data object's duplicate tweet list.

The above three stages in eliminating duplicates in processed tweets results around 90% of the original tweets to be discarded.

*D. Parallel Thread Operation*

Although the algorithm described above was giving the correct logical output, it was underutilizing the computational power of the computes that it was run owing to the sequential manner the program was written so far. It was also observed that each of the parts that took the most computation time; Class duplicate handling and duplicate eliminating was actually consisted of for-loops where each iteration was completely independent of the previous iterations. Thus it was decided to parallelize both operations. After extensive analysis of the current algorithms used; it was decided that an architecture where a central master thread communicate with satellite threads via events is the best possible threaded architecture for this system. It was also observed that the duplicate eliminating process has to be started only after the class duplicate handling process was over. Thus a thread barrier was needed to be implemented.

Since Java does not have an inbuilt threaded event handling system it was decided to use the free java package called LinkSet [5] as the base of the event architecture. By extending the java.lang.Thread class while implementing constructs of the LinkSet package, a class called "LockingThread" was created along with a thread barrier. Fig. 3 shows the pseudocode for the Barrier Algorithm in the LockingThread. The class duplicate handling methods were ported to a new class called "ClassDuplicateHandler" which extended the LockingThread class. Fig. 4 shows the pseudocode for the class duplicate handler threading algorithm. Similarly the "Duplicate eliminating" methods were ported to a new class called "DuplicatesEliminator" which also extended the "LockingThread" class. Fig. 5 show the pseudocode for the duplicate eliminator threading algorithm.

"ClassDuplicateHandler" class and the "DuplicatesEliminator" class were to be the satellite thread instances. As for the master thread; the Generator class was altered by making it a child of the "LockingThread" class and introducing a variable length to the thread barrier. The length of the barrier indicates the number of threads that are executable in parallel.

```
Declare comp := boolean[]

LOCK( )
  while !EVALUATE-COMPONENTS( )
    do sleep(500)
  ResetComponents( )

RESETCOMPONENTS( )
  for i ← 0 to length[comp]
    do comp[i] ← false

boolean EVALUATE-COMPONENTS( )
  for i ← 0 to length[comp]
    do if comp[i] ≠ true
      then return false
  return true
```

Fig. 3. Barrier Algorithm

```
for i ← 1 to length[A] / 2
  do for j ← 1 to length[comp]
    do threshold ← (length[A] / 2) + (length[A] % 2)
      if length[A] - (i+j) > threshold
        then Thread t ← CLASSDUPLICATEHANDLER(A[i+j], A[length[A] - (i + 1 + j)), j, generator)
             t.start( );
      else
        then for k ← j to length[comp]
          do comp[k] ← true
          break
      j ← j+1
  lock( )
  i ← i + length[comp]
```

Fig. 4. Class Duplicate Handler threading algorithm

```
for j ← 1 to length[B]
  do for i ← 1 to length[comp]
    do if (i+j) < length[B]
      then Thread t ← DUPLICATESELIMINATOR(B[i+j], B, i + j, i, generator)
           t.start()
    else
      then comp[i] ← true
    i ← i+1
  LOCK()
  j ← j + length[comp]
```

Fig. 5. Duplicate eliminator threading algorithm

The program was run on two computers with Intel® Core™ Duo @2.10GHz, 2.10GHz processors. Thus the thread count was set to four to be twice the number of CPUs involved. This software configuration reported around 38% decrease in execution time from the sequential program. This is highly significant considering the fact that in the given hardware configuration it takes around 38 minutes to process around 300,000 tweets by the sequential program.

*E. Building Trained Data Model*

The final processed data was used in 3 different classifiers to identify which classifier provides a reasonably accurate data model for tweet prediction purposes. Data mining tool Weka, developed by University of Waikato[1] was used for this purpose as it has number of classifiers developed for similar purposes. The newest version of Weka, 3.7.2 was used as per recommendation from its developers. Naïve Bayes, Random Tree and Random Forest were chosen to develop the data model from the processed training data set.

*1) Naive Bayes Classifier:* The Naive Bayes Classifier technique is based on the Bayesian theorem and is particularly suited when the dimensionality of the inputs is high [6]. This classifier assumes that the presence or absence of a particular attribute does not affect the presence or absence of other attributes.

*2) Random Tree Classifier:* Randome tree algorithm constructs a tree that considers K random features (Smaller than the total number of features) from the data set at each node. It does not perform pruning.

*3) Random Forest Classifier:* The random forest algorithm creates an ensemble of classifiers by training each classifier on a random redistribution of the training set. Each random redistribution is generated by randomly drawing with replacement *N* examples where *N* is the size of the training set. A tree is grown on a fixed-size subset of attributes (smaller than the total number of attributes) randomly drawn on each round [7].

The results indicated very high accuracy obtained through Random Forest and Random tree classifiers where as naïve Bayes classifier produced poor inaccurate model. Due to high accuracy of using Random Forest classifier against test data, it was chosen as the classifier to be used against proposed data preprocessing system.

## III. RESULTS AND ANALYSIS

The train data collected was used in number of classifiers to build the data model. The following are the results for three classifiers Naïve Bayes, Random Tree and Random Forest using 17, 861 train data items and 100 sample test data items.

TABLE I
CLASSIFIER TEST RESULTS

|  | **Naïve Bayes** | **Random Tree** | **Random Forest** |
|---|---|---|---|
| Trained model Accuracy | 15.62% | 100% | 99.524% |
| Trained model kappa statistic | 0.0289 | 1 | 0.9936 |
| Average test data accuracy | 23% | 52% | 56% |

Though random Tree model shows highest accuracy over trained model, since it shows a low accuracy in terms of test data than Random Forest model, it was discarded as a suitable model. Naïve Bayes showed poor performance against train and test data where as Random Forest model showed significantly high accuracy over train data.

The following demonstrates results of the experiment carried out on three samples of test data containing 100 data items each on Random Forest classifier.

TABLE II
RANDOM FOREST CLASSIFIER TEST RESULTS

| **Sample** | **Accuracy (%)** |
|---|---|
| Sample 1 | 58 |
| Sample 2 | 56 |
| Sample 3 | 55 |

The average sample test data set accuracy shows 56% in correctly classifying the tweets. Considering the fact that the probability of correctly classifying a tweet manually being 20%, model shows an accuracy increase of 36%.

TABLE III
RANDOM FOREST CLASSIFIER CONFUSION MATRIX

| **a** | **b** | **c** | **d** | **e** | ←**Classified as** | |
|---|---|---|---|---|---|---|
| 14 | 3 | 2 | 0 | 2 | **a** | **Business** |
| 2 | 10 | 1 | 1 | 1 | **b** | **Politics** |
| 4 | 2 | 8 | 3 | 2 | **c** | **Technology** |
| 2 | 3 | 3 | 12 | 8 | **d** | **lifestyle** |
| 4 | 1 | 0 | 0 | 12 | **e** | **art** |

TABLE III shows the confusion matrix for test data sample 2 with 56% overall accuracy. As the matrix shows, technology and lifestyle classes are misclassified more than 50% of the time resulting reduction in overall accuracy. This is due to limited availability of tweet data of the above two classes in train data.

Parallel thread operation on duplicate elimination algorithm provides the following results on processing 307,000 tweets to 17,861 processed data items.

TABLE IV
PARALLEL THREAD OPERATION RESULTS

| | |
|---|---|
| Data Processing with no threads | 421 minutes |
| Data Processing with 4 threads | 259 minutes |
| Execution time speed up | 38.48% |

## IV. FURTHER IMPROVEMENTS

The Data processing could be further improved through substituting a URL with the actual web site content as in heading of the web page. This would greatly improve the classification and identification of words in to attributes identified through indexing algorithm. In addition, use of significantly high number tweets (greater than 2 million) would improve the accuracy of the data model thus resulting higher accuracy in test data.

The data processing proposed is limited to the word count in WordNet library. Thus any improvement in the WordNet library or use of another superior word library would extract information more from the tweets in to attributes thus the processed tweet would be better represented through the attribute values.

## V. CONCLUSION

Twitter data stream allows users to get real time updates on the global sentiment. Due to the omnifariousness of the subjects in the stream, it is highly inefficient to run a data mining algorithm on the raw data. This paper discussed an algorithm to classify the stream in to a given number of mutually exclusive collectively exhaustive streams using the word repository of WordNet. Weka was used to observe the accuracy of the selected attribute combinations. Considering all tests performed and the subsequent results obtained; it can be concluded that the suggested algorithm is in fact suitable for selecting the most relevant attributes and building the train data set for the stream classifying operation.

## APPENDIX

### A. An excerpt from the noun.person file.

| |
|---|
| { Zeus, Greek_deity,@i noun.group:Greek_mythology,;c ((Greek mythology) the supreme god of ancient Greek mythology; son of Rhea and Cronus whom he dethroned; husband and brother of Hera; brother of Poseidon and Hades; father of many gods; counterpart of Roman Jupiter) } |
| { Hera, Here, Greek_deity,@i (queen of the Olympian gods in ancient Greek mythology; sister and wife of Zeus remembered for her jealously of the many mortal women Zeus fell in love with; identified with Roman Juno) } |
| { Greek_deity, deity,@ noun.time:antiquity,;c (a deity worshipped by the ancient Greeks) } |
| { [ deity, verb.cognition:deify,+ ] [ divinity, adj.all:heavenly^divine2,+ ] god1, immortal, supernatural_being,@ noun.group:pantheon,#m (any supernatural being worshipped as controlling some part of the world or some aspect of life or who is the personification of a force) } |
| { [ God, adj.all:heavenly^godly,+ ] Supreme_Being, supernatural_being,@i (the supernatural being conceived as the perfect and omnipotent and omniscient originator and ruler of the universe; the object of worship in monotheistic religions) } |
| { self, noun.Tops:person,@ (a person considered as a unique individual; "one's own self") } |
| { Parkinson1, James_Parkinson, surgeon,@i (English surgeon (1755-1824)) } |
| { Morpheus, deity,@i Ovid,;c (the Roman god of sleep and dreams) } |
| { democrat, [ populist, noun.cognition:populism,+ ] advocate,@ (an advocate of democratic principles) } |
| { Dalai_Lama, Grand_Lama, lama,@ (chief lama and once ruler of Tibet) } |
| { dame1, madam, ma'am, lady1, gentlewoman, woman,@ (a woman of refinement; "a chauffeur opened the door of the limousine for the grand lady") } |
| { countryman, compatriot,@ (a man from your own country) } |
| { Otto_I, Otho_I, Otto_the_Great, King_of_the_Germans,@i Holy_Roman_Emperor,@i (King of the Germans and Holy Roman Emperor (912-973)) } |
| { paper-pusher, bureaucrat,@ clerk,@ (a clerk or bureaucrat who does paperwork) } |
| { outdoorsman, noun.Tops:person,@ (a person who spends time outdoors (e.g., hunting or fishing)) } |
| { clog_dancer, dancer1,@ (someone who does clog dancing) } |
| { cowgirl, cowboy,@ (a woman cowboy) } |
| { church_officer, official1,@ (a church official) } |
| { [ butcher2, verb.contact:butcher,+ ] meatman, merchant,@ (a retailer of meat) } |
| { bounty_hunter1, pursuer,@ (someone who pursues fugitives or criminals for whom a reward is offered) } |
| { bird_fancier, fancier,@ (a person with a strong interest in birds) } |

6ACKNOWLEDGMENT

While doing this research we had to go on a long journey, meeting many challenges and overcoming those with contribution of the people who surrounded us with helping hand and courageous words. We would like to thank our subject coordinator Dr. Shehan Perera for the support and encouragement given in regard to this project.

REFERENCES

[1] University of Waikato. (2010). Weka [Online]. Available: www.cs.waikato.ac.nz/ml/**weka/**
[2] Mix Online Production. (2010). tHE ARCHIVISt desktop[Online]. Available: http://visitmix.com/labs/archivist-desktop/
[3] Mix Online Production. (2010). tHE ARCHIVISt [Online]. Available: http://archivist.visitmix.com/
[4] Princeton University. (2010) WordNet – A Lexical Database for English [Online]. Available: http://wordnet.princeton.edu/
[5] Łukasz Bownik. (2010, February 19). LinkSet – An alternative Approach to Events Java [Online]. Available: http://www.codeproject.com/KB/library/Linkset.aspx
[6] Wikipedia. (2011, February 1). Naïve Bayes Classifier [Online]. Available: http://en.wikipedia.org/wiki/Naive_Bayes_classifier
[7] Wikipedia. (2011, February 22). Random Forest [Online]. Available: http://en.wikipedia.org/wiki/Random_forest